\documentclass[11pt]{article}
\usepackage[utf8]{inputenc}

\usepackage{acl}
\usepackage{booktabs}
\usepackage{times}
\usepackage{latexsym}

\usepackage[T1]{fontenc}

\usepackage[utf8]{inputenc}

\usepackage{microtype}

\usepackage{inconsolata}

\usepackage{graphicx}

\definecolor{codebg}{RGB}{245,245,245}
\definecolor{stringcol}{RGB}{163,21,21}
\definecolor{keycol}{RGB}{0,0,200}
\definecolor{commentcol}{RGB}{106,153,85}

\usepackage{booktabs}
\usepackage{tabularx}
\usepackage{listings}
\usepackage{xcolor}

\definecolor{jsonbg}{RGB}{248,248,248}
\definecolor{jsonstring}{RGB}{163,21,21}

\lstdefinelanguage{json}{
    basicstyle=\ttfamily\small,
    showstringspaces=false,
    breaklines=true,
    frame=single,
    rulecolor=\color{gray!40},
    backgroundcolor=\color{jsonbg},
    string=[s]{"}{"},
    stringstyle=\color{jsonstring},
    comment=[l]{//},
    morecomment=[s]{/*}{*/},
}

\usepackage{authblk}
\usepackage{hyperref} 

\makeatletter
\renewcommand\AB@authnote[1]{} 
\renewcommand\AB@affilnote[1]{}
\makeatother

\title{\textbf{BharatGather: A Culturally-Informed Benchmark Dataset for Misinformation and Fake News Detection in Indian Public Events
}}

\author{%
  \textbf{Parth Bramhecha} \enspace
  \textbf{Smit Deshmukh} \enspace
  \textbf{Sairaj Bodhale} \enspace
  \textbf{Adwait Borate} \enspace
  \textbf{Raviraj Joshi} \\
  \vspace{3pt}
  \textit{L3Cube-Labs, Pune} \\
  \vspace{3pt}
  \small\texttt{\{}\href{mailto:parth.bramhecha007@gmail.com}{\texttt{parth.bramhecha007}}\texttt{,}
  \href{mailto:deshmukhsmit11@gmail.com}{\texttt{deshmukhsmit11}}\texttt{,}
  \href{mailto:adwaitborate@gmail.com}{\texttt{sairaj.sab}}\texttt{,}
  \href{mailto:sairaj.sab@gmail.com}{\texttt{adwaitborate}}\texttt{,}
  \href{mailto:raviraj.j1991@gmail.com}{\texttt{ravirajoshi}}\texttt{\}@gmail.com}
}

\date{}

\begin{document}
\maketitle
\begin{abstract}
Large-scale public events, such as religious festivals, political rallies, and cultural gatherings, are increasingly vulnerable to the rapid dissemination of misinformation, posing substantial risks to public safety and social cohesion. While automated fake news detection has seen significant methodological progress, existing benchmarks frequently fail to capture the socio-cultural nuances and event-specific dynamics characteristic of the Indian context. This paper introduces BharatGather, a curated, multi-source dataset specifically engineered for binary misinformation classification within the ecosystem of Indian mass gatherings. The corpus comprises 14,646 records constructed through a hybrid pipeline involving systematic web scraping of prominent fact-checking platforms, multimedia transcript extraction, and Large Language Model (LLM)-mediated synthetic augmentation to ensure narrative diversity. By providing a resource tailored to the unique complexities of event-aware misinformation in India, this work facilitates the development of culturally informed detection systems and establishes a rigorous benchmark for evaluating their performance in high-stakes public environments.
\end{abstract}

\section{Introduction}
The digital proliferation of misinformation has emerged as a formidable societal challenge, with its impact being most pronounced in high-density public environments\cite{r3,r11}. In the Indian socio-political landscape, large-scale gatherings ranging from religious pilgrimages to political demonstrations function as epicenters for intense information exchange across social media, broadcast news, and video-sharing platforms\cite{r17}. Within these volatile information ecosystems, the rapid diffusion of false or misleading claims can catalyze panic, communal friction, and hazardous crowd dynamics\cite{r16,r20}. Current research in Natural Language Processing (NLP), bolstered by the evolution of transformer-based architectures, has significantly enhanced the efficacy of automated veracity assessment\cite{r14,r19,wani2021evaluating}. However, a systemic limitation persists: the majority of existing research utilizes domain-general or Western-centric datasets that lack the cultural and linguistic granularity required for effective deployment in regional contexts\cite{r8,r20}. Furthermore, while some studies have addressed the Indian landscape, they often focus on architectural frameworks without releasing the underlying structured datasets essential for reproducible research in event-driven misinformation\cite{r17}.

To address these deficits, we introduce \textbf{BharatGather}\footnote{\href{https://huggingface.co/datasets/l3cube-pune/BharatGather}{BharatGather Dataset} (l3cube-pune/BharatGather)}, a curated dataset designed for \textbf{binary misinformation detection} (categorized into True and False labels) within the specific context of Indian mass gatherings\cite{r14}. Our contribution integrates diverse data acquisition modalities, utilizing web-scraped news corpora, extracted transcripts from multimedia platforms, and controlled synthetic augmentation via LLMs to enhance the diversity of the misinformation narratives\cite{r7,r8}. The final dataset comprises \textbf{14,646 records}, each enriched with veracity labels and event-aware metadata to support systematic benchmarking\cite{r15}.

The primary contributions of this work are focused on the development and validation of this specialized benchmark. First, we present a culturally-informed corpus of 14,646 records centered on regionally significant mass gatherings in India, filling a distinct gap in the current literature regarding event-specific datasets\cite{r20}. This was achieved through a multi-tier acquisition pipeline that integrated structured data from five leading fact-checking organizations—AltNews, BoomLive, Factly, The Quint, and NewsMeter—alongside YouTube news transcripts to capture the multimedia dimension of modern misinformation\cite{r5,r17}. 

Furthermore, this research implements a sophisticated adversarial narrative generation strategy. Using Qwen3-32B, we produced 3,595 synthetic variants designed to expose models to subtle deceptive patterns that exceed surface-level lexical analysis\cite{r7,r8}. Finally, we establish a rigorous empirical baseline for binary classification using \texttt{bert-base-uncased}\cite{r8,r19}. Our evaluations demonstrate that while the dataset contains a robust learnable signal, achieving 96.1\% accuracy, the significant performance delta between frozen and fine-tuned models underscores the necessity of domain-specific gradient adaptation for high-fidelity detection\cite{r8,r19}. Collectively, these contributions provide a foundation for automated misinformation mitigation within high-stakes, high-density public environments in India\cite{r13}.

\section{Related Work}

Research on misinformation detection spans computational, behavioral, and communication perspectives. Foundational work highlighted the socio-technical nature of online misinformation and the need for interdisciplinary interventions \cite{r2}. Broad surveys further established the effectiveness of transformer-based approaches while emphasizing persistent limitations in data quality, contextual bias, and domain transfer \cite{r8}.

\subsection{Misinformation in the Indian Context}

Prior studies indicate that misinformation in India is shaped by distinct cultural, political, and religious dynamics. NLP-based misinformation matching systems have demonstrated practical feasibility in Indian settings, but large-scale event-specific benchmarks remain scarce \cite{r1}. Thematic analyses of misinformation during COVID-19 in India report recurring narratives around health claims, political manipulation, religious polarization, and communal framing \cite{r10,r17}. This evidence supports the need for culturally grounded dataset construction.

\subsection{Detection and Propagation Modeling}

Beyond text classification, several works model temporal and network effects in misinformation spread. Empirical studies show that early factual interventions can reduce subsequent misinformation diffusion \cite{r4}, while graph-based formulations identify influential users as key amplification or mitigation agents \cite{r16}. Additional methods use statement-level semantic conflict and behavioral forensics to improve misinformation identification performance \cite{r14,r19}.

\subsection{Crowd-Based and Mitigation Approaches}

Crowd and intervention methods provide complementary mitigation pathways. Recent surveys classify crowd roles as annotators, evaluators, and creators \cite{r5}. Synthetic crowd modeling has been proposed to approximate representative veracity judgments \cite{r7}. Other approaches include reinforcement learning for staged debunker selection and prebunking strategies for proactive suppression \cite{r12,r13}. Evidence from field experiments in India also suggests that intervention effects vary across audience groups and political alignment \cite{r6}.

\subsection{Crisis Communication and Information Behavior}

From a crisis communication viewpoint, prior work documents operational gaps in monitoring and response workflows \cite{r3}. Information behavior frameworks model disinformation across creation, dissemination, and consumption stages \cite{r11,r15}. Platform-level vulnerabilities and sustained persuasion effects from fake news exposure further motivate timely and context-sensitive detection systems \cite{r9,r18}.

\section{Dataset Construction}

\begin{figure*}[t]
    \centering
    \includegraphics[width=\textwidth]{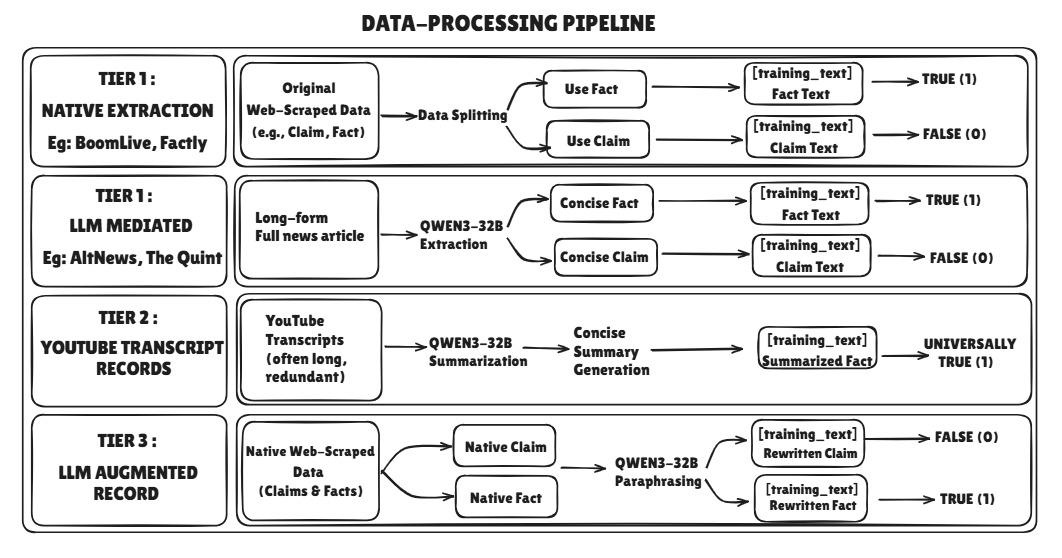}
    \caption{Overview of dataset construction}
    \label{fig:dataset_construction}
\end{figure*}

\subsection{Dataset Scope}
This corpus is specifically engineered to capture misinformation within the ecosystem of Indian mass gatherings, including religious festivals, political rallies, and regional cultural events such as the Kumbh Mela, Ganesh Chaturthi, Durga Puja, Farmers' Protest, Rath Yatra, Sabarimala Pilgrimage, and Maratha Kranti Morcha. It reflects an evolving socio-digital paradigm in India, where deceptive narratives frequently function as a \textit{primary catalyst} for crowd mobilization rather than emerging as a secondary consequence. Every instance is anchored in verified reporting from established Indian fact-checking organizations, ensuring domain-specific relevance and mitigating the noise characteristic of generic global datasets.

\subsection{Data Sources}

To ensure representational diversity and mitigate corpus homogeneity, data were aggregated from three distinct modalities:

\textbf{Established Fact-Checking Platforms.} Five prominent Indian
fact-checking organizations serve as the primary structured sources:
\textit{AltNews}, \textit{BoomLive}, \textit{Factly}, \textit{The
Quint}, and \textit{NewsMeter}. These platforms routinely publish
verified claims alongside corresponding fact-checked responses, making
them high-integrity foundations for ground-truth labeling.

\textbf{YouTube News Transcripts.} Transcripts were extracted from authenticated and credible Indian news channels on YouTube, including sources such as \textit{BBC News} and \textit{NDTV}. This modality captures the multimedia dimension of contemporary news consumption and reflects how misinformation propagates through video-based platforms during mass events.

\textbf{Synthetically Augmented Records.} To bolster model robustness
and address class-distribution concerns, controlled synthetic records
were generated using large language models conditioned on high-integrity
verified instances.

\subsection{Data Acquisition Pipeline}

Data collection was structured as a three-tier acquisition and
processing pipeline, each tier differentiated by source type, structural
integrity, and downstream processing logic.

\subsubsection{Tier I: Web-Scraped Structured Data}

The core corpus was assembled via systematic web scraping of the identified platforms. Custom scrapers were developed to accommodate heterogeneous HTML architectures, yielding \textbf{10,691 records}.

\paragraph{Native Entity Extraction:} 
For BoomLive, Factly, and NewsMeter, \texttt{Fact} and \texttt{Claim} fields were available as discrete metadata. These \textbf{2,368 records} constitute the highest-integrity tier and served as the exclusive basis for downstream synthetic augmentation.
\\In this context, a \textbf{Claim} represents a widely circulating piece of an unverified assertion found on social media. The \textbf{Fact} represents the objective, verified ground truth established by professional journalists after rigorous investigation.

\paragraph{LLM-Mediated Semantic Extraction:}
AltNews and The Quint provided predominantly unstructured raw text
fields without pre-segmented fact-claim pairs. To normalize these
\textbf{8,323 records} into the unified schema, we employed
\textit{Qwen3-32B}  to extract the
underlying \texttt{Fact} and \texttt{Claim} from each article. The
model was prompted to identify the semantic crux of each piece,
condensing the raw text into concise, schema-compatible representations.
To preserve data integrity and prevent recursive model bias, commonly
referred to as \textit{model collapse}, these
LLM-derived records were deliberately excluded from the synthetic
augmentation stage, maintaining a clear provenance boundary in the
training corpus. The prompt schema used for this extraction is shown in
Listing~\ref{lst:extraction_prompt}.

\begin{lstlisting}[language=json,
    caption={Prompt schema for LLM-mediated fact-claim extraction
             (AltNews \& The Quint).},
    label={lst:extraction_prompt}]
{
  "role": "system",
  "instruction": "You are a fact-checking assistant. Given a raw
    news article, extract the two fields defined in output_schema.",
  "output_schema": {
    "Claim": "The specific assertion in the article that requires
              verification.",
    "Fact":  "The verified ground truth that supports or contradicts
              the claim."
  },
  "constraints": [
    "Do not add information absent from the source text.",
    "Keep each field concise (1-3 sentences).",
    "Preserve names, dates, and numerical values verbatim.",
    "Return only valid JSON with keys 'Claim' and 'Fact'."
  ],
  "model":       "qwen/qwen3-32b",
  "temperature": 0.6,
  "max_tokens":  4096,
  "top_p":       0.95
}
\end{lstlisting}

\paragraph{Platform Variations and Structural Diversity}
While all five platforms serve as primary debunking sources, their digital presentation of information varies significantly, necessitated the specialized extraction methods detailed in Tier I:

\textbf{Structured Layouts (BoomLive, Factly, NewsMeter):} These websites utilize a "Fact-Check Card" format. They explicitly segregate the viral claim from the final verdict, often using specific HTML tags like \texttt{claim-review} or \texttt{fact-check-summary} This allows for direct native extraction of the fact-claim pairs.
    
\textbf{Narrative Layouts (AltNews, The Quint):} These platforms often employ a long-form investigative journalism style. The misinformation and the debunking evidence are interwoven throughout the article body rather than being isolated in metadata fields. Consequently, LLM-mediated extraction via was required to distill these narratives into our standardized schema.

\subsubsection{Tier II: YouTube Transcript Integration}

To capture the video-mediated dimension of misinformation, transcripts
were extracted from verified Indian news channels on YouTube. A
deliberate selection criterion restricted sourcing to established,
credible news organizations to minimize the introduction of noisy or
unverified content. This tier contributed \textbf{360 records} to the
corpus. A known limitation of this source is inter-channel redundancy:
the same news event is frequently reported across multiple channels,
resulting in near-duplicate transcripts. Deduplication heuristics were
applied to mitigate this effect, though some thematic overlap may
persist.

\subsubsection{Tier III: LLM-Based Synthetic Augmentation}

To enhance representational diversity and address potential class
imbalance, synthetic records were generated exclusively from the
highest-integrity Tier~I records (BoomLive, Factly, and NewsMeter). The
pipeline was implemented using \textit{Qwen3-32B}. A lexical overlap score was computed after each generation using \texttt{difflib.SequenceMatcher}, enforcing a bounded
similarity range with up to three retries per article. For each seed
record the pipeline produced two variants.

\paragraph{True Variant Generation.}
A discourse style was sampled uniformly at random from the set
\{\textit{formal news report, investigative journalism style, feature
article style, bulletin brief, analytical piece}\}. The model was
prompted to produce a factually identical paraphrase of the source in
the selected style. The prompt schema is given in
Listing~\ref{lst:true_prompt}.

\begin{lstlisting}[language=json,
    caption={Prompt schema for true-variant synthetic generation.},
    label={lst:true_prompt}]
{
  "role": "You are rewriting a verified news article.",
  "goal": "Produce a factually identical article in {discourse_style}.",
  "discourse_styles": [
    "formal news report",
    "investigative journalism style",
    "feature article style",
    "bulletin brief",
    "analytical piece"
  ],
  "requirements": [
    "Preserve every factual element.",
    "Preserve names, dates, and numbers exactly.",
    "Deep paraphrasing required --- do not copy sentences.",
    "Neutral tone throughout.",
    "No added interpretation or extra facts."
  ],
  "output_format": {
    "HEADLINE": "<rewritten headline>",
    "TEXT":     "<rewritten article body>"
  },
  "generation_config": {
    "model":       "qwen/qwen3-32b",
    "temperature": 0.6,
    "max_tokens":  4096,
    "top_p":       0.95
  },
  "quality_gate": {
    "lexical_overlap_min": 0.60,
    "lexical_overlap_max": 0.90,
    "max_retries":         3
  }
}
\end{lstlisting}

\paragraph{False Variant Generation.}
For adversarial variants, two to three distortion strategies were
sampled without replacement from a predefined distortion taxonomy. The
model was instructed to apply the selected distortions while retaining
the same actors, event, and overall narrative structure. The lexical
overlap threshold was set to $(0.65,\,0.92)$, reflecting the need for
the misleading text to remain semantically proximate to the source. The
prompt schema is given in Listing~\ref{lst:false_prompt}.

\begin{lstlisting}[language=json,
    caption={Prompt schema for false-variant (adversarial) synthetic
             generation.},
    label={lst:false_prompt}]
{
  "role": "Generate a subtly misleading article for adversarial
           misinformation detection research.",
  "rules": [
    "Keep the same people and event as the source.",
    "Introduce no new actors.",
    "Make no absurd or obviously false claims.",
    "Maintain a neutral, professional tone.",
    "Apply exactly 2-3 of the sampled distortions.",
    "Maintain high semantic similarity to the source."
  ],
  "distortion_taxonomy": [
    "Imply stronger conclusion than supported",
    "Remove uncertainty qualifier",
    "Exaggerate numbers slightly",
    "Change specific dates to vague times",
    "Shift the blame or responsibility subtly",
    "Omit crucial context",
    "Use marginally stronger emotional language"
  ],
  "distortions_applied": "{2_or_3_sampled_from_taxonomy}",
  "output_format": {
    "HEADLINE": "<misleading headline>",
    "TEXT":     "<subtly distorted article body>"
  },
  "generation_config": {
    "model":       "qwen/qwen3-32b",
    "temperature": 0.6,
    "max_tokens":  4096,
    "top_p":       0.95
  },
  "quality_gate": {
    "lexical_overlap_min": 0.65,
    "lexical_overlap_max": 0.92,
    "max_retries":         3
  }
}
\end{lstlisting}

This augmentation strategy yielded an additional \textbf{3,595 records}
and exposes downstream classifiers to subtle deceptive patterns that are
difficult to detect through surface-level lexical analysis alone.

\subsection{Data Labeling}

Records were labeled according to a strict directional semantic matching
protocol. If the extracted \texttt{Claim} reached semantic alignment
with the verified \texttt{Fact}, the instance was assigned a binary
label of \texttt{1} (True). Conversely, if the claim contradicted or
materially diverged from the verified fact, the instance was labeled
\texttt{0} (False). For synthetically generated pairs, labels were
assigned deterministically based on the generation condition:
true-hard variants received label \texttt{1} and false-hard variants
received label \texttt{0}.

\subsubsection{Training Text Generation}

The \texttt{training\_text} field serves as the primary input string to all downstream NLP models and is constructed through distinct curation logic tailored to the structural characteristics of each acquisition tier.

For Tier~I LLM-mediated records (AltNews and The Quint), where the source material consists of long-form articles, \textit{Qwen3-32B} is used to extract a concise \texttt{Fact} and \texttt{Claim} from the narrative. Each article is then split into two separate records: the extracted \texttt{Fact} is used as the \texttt{training\_text} with a label of true (1), while the extracted \texttt{Claim} is used with a label of false (0). This distillation removes unnecessary rhetoric and ensures the model trains on the core semantic meaning of the investigative report.

For Tier~II YouTube transcript records, the \texttt{training\_text} is created by summarizing raw video transcripts. Since transcripts are often long and repetitive, \textit{Qwen3-32B} is used to generate a concise summary of the core content. Because these transcripts represent verified information, the resulting summary is used as the \texttt{training\_text} and is universally labeled as true (1). This ensures the model trains on the essential facts of the video without the noise of conversational speech.

For Tier~III LLM-augmented records, we expand the dataset by using an LLM to paraphrase our original web-scraped data. From each native record, we generate two distinct training samples: the original \texttt{Claim} is rewritten to create a new claim record labeled false (0) which acts as \texttt{training\_text}, and the original \texttt{Fact} is rewritten to create a fact record labeled true (1) which acts as \texttt{training\_text}. This ensures that the model learns to identify factual consistency across different phrasing styles while doubling our training examples.

\subsection{Dataset Schema}

To ensure architectural consistency across all acquisition tiers, every
record in the final corpus is mapped to a standardized ten-attribute
schema. This structured format preserves both the provenance and the
linguistic context of each instance. A representative record is shown in
Listing~\ref{lst:schema}.

\begin{lstlisting}[language=json,
    caption={Universal dataset schema — one representative record.},
    label={lst:schema}]
{
  "id":           "GE_1042",
  "headline":     "Officials confirm festival crowd management
                   plan activated",
  "src":          "BoomLive",
  "label":        1,
  "training_text":"<primary NLP input string curated per tier>",
  "full_article": "<complete source text or transcript>",
  "url":          "https://www.boomlive.in/...",
  "content_type": "fact_checked"
}
\end{lstlisting}

The eight fields carry the following semantics. \textbf{id} is a unique
auto-incremented identifier for record tracking and retrieval.
\textbf{headline} is the original or rewritten title of the news item.
\textbf{src} identifies the originating platform. \textbf{label} is a binary indicator where \texttt{1} denotes verified truth and \texttt{0} denotes misinformation.
\textbf{training\_text} is the primary input string for the NLP model,
curated according to the processing logic of each acquisition tier. \textbf{full\_article} is the complete source text or transcript, providing exhaustive context beyond the training string. \textbf{url} is the direct hyperlink to the source
for independent verification and auditability. \textbf{content\_type}
distinguishes between \texttt{fact\_checked} records (natively extracted
dual-attribute sources) and \texttt{factual\_data} records (LLM-derived
or single-source summaries); synthetically generated records carry the
value \texttt{"Generated"}.

\subsection{Dataset Statistics}

The final dataset comprises \textbf{14,646 records} distributed across
the three acquisition tiers. Table~\ref{tab:tier_distribution}
summarizes the tier-level distribution, and
Table~\ref{tab:source_distribution} provides the per-platform breakdown
for Tier~I.

\begin{table}[h]
\centering
\caption{Record distribution by data acquisition tier.}
\label{tab:tier_distribution}
\renewcommand{\arraystretch}{1.25}
\begin{tabularx}{\columnwidth}{lXr}
\toprule
\textbf{Tier} & \textbf{Source Type} & \textbf{Records} \\
\midrule
Tier I   & Web Scraping (Structured)             & 10,691 \\
Tier II  & YouTube Transcripts (Semi-Structured) & 360    \\
Tier III & Synthetic Augmentation (LLM)          & 3,595  \\
\midrule
\textbf{Total} & & \textbf{14,646} \\
\bottomrule
\end{tabularx}
\end{table}

\begin{table}[h]
\centering
\caption{Per-platform breakdown of Tier~I web-scraped records.}
\label{tab:source_distribution}
\renewcommand{\arraystretch}{1.25}
\begin{tabularx}{\columnwidth}{lXrr}
\toprule
\textbf{Platform} & \textbf{Extraction Method} &
\textbf{Records} & \textbf{Synthetic} \\
\midrule
AltNews   & LLM Extraction    & 3,693 & No  \\
The Quint & LLM Extraction    & 4,630 & No  \\
BoomLive  & Native Extraction & 698   & Yes \\
Factly    & Native Extraction & 792   & Yes \\
NewsMeter & Native Extraction & 878   & Yes \\
\midrule
\textbf{Total} & & \textbf{10,691} & \\
\bottomrule
\end{tabularx}
\end{table}


\section{Experiments}

\subsection{Dataset Overview}

\begin{figure}[h]
  \centering
  \includegraphics[width=\columnwidth]{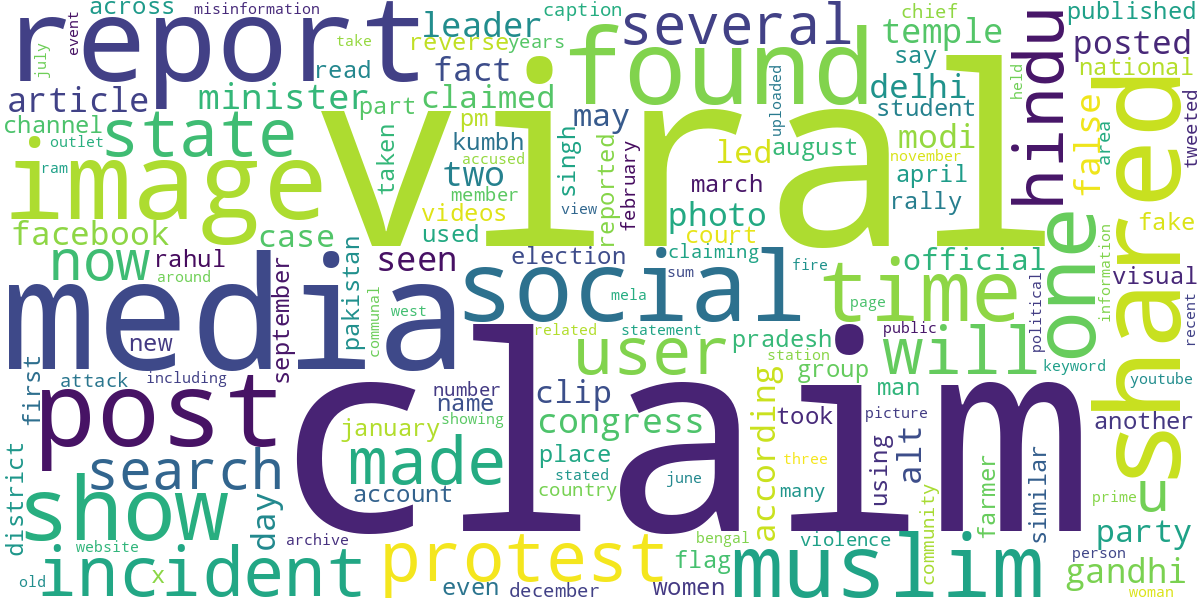}
  \caption{Word cloud of the \texttt{training\_text} field. Font size is proportional
           to term frequency.}
  \label{fig:wordcloud}
\end{figure}

Figure~\ref{fig:wordcloud} presents a word cloud derived from the
\texttt{training\_text} field of the corpus.  Dominant terms such as
\emph{viral}, \emph{claim}, \emph{media}, \emph{report}, and
\emph{post} reflect the social-media provenance of the data, while
politically and religiously charged vocabulary (\emph{hindu},
\emph{muslim}, \emph{congress}, \emph{modi}) underscores the
domain-specific challenges of Indian-context misinformation detection.

\subsection{Experimental Setup}

The classification experiments were conducted using
\texttt{bert-base-uncased} as the architectural backbone, fine-tuned for
binary sequence classification (\texttt{label~1} = True,
\texttt{label~0} = Misinformation).  Prior to partitioning, the corpus
was subjected to rigorous de-duplication based on the
\texttt{training\_text} field 

Table~\ref{tab:dataset_split} summarises the resulting dataset splits
used across all experiments.

\begin{table}[h]
\centering
\caption{Dataset split statistics after de-duplication}
\label{tab:dataset_split}
\small
\setlength{\tabcolsep}{6pt}
\renewcommand{\arraystretch}{1.05}
\resizebox{\columnwidth}{!}{%
\begin{tabular}{@{}lrrr@{}}
\toprule
\textbf{Configuration} & \textbf{Train} & \textbf{Validation} & \textbf{Test} \\
\midrule
BERT Frozen & 7{,}775 & 864 & 2{,}149 \\
BERT Fine-tuning & 7{,}442 & 827 & 2{,}078 \\
\bottomrule
\end{tabular}%
}
\end{table}

Model initialisation was performed using pretrained
\texttt{bert-base-uncased} weights.  Two primary training regimes were
evaluated to establish baseline performance:

\begin{enumerate}
  \item \textbf{Feature Extraction (Frozen):}  The entire BERT backbone
        was frozen and only the newly initialised linear classification
        head was trained.  Class-imbalance was addressed via weighted
        cross-entropy loss (weights: 1.20 for class~0,~0.86 for
        class~1).  Training ran for 5~epochs.

  \item \textbf{Bert Fine-tuning :}  All encoder parameters across all transformer layers were updated in conjunction with the classification
head, allowing for global adaptation to the domain-specific nuances of the misinformation dataset. Training ran for 3~epochs.
\end{enumerate}

All experiments used a maximum sequence length of 256~tokens, a
per-device batch size of 16, a learning rate of $3\times10^{-5}$, and a
weight decay of 0.01.  Optimisation was governed by validation macro
F1, and the best-performing checkpoint was selected for final
evaluation.

\subsection{Results}

Table~\ref{tab:results} compares test-set performance across both
configurations.

\begin{table}[h]
\centering
\caption{Test-set performance across BERT training configurations.}
\label{tab:results}
\small
\setlength{\tabcolsep}{4pt}
\renewcommand{\arraystretch}{1.05}
\resizebox{\columnwidth}{!}{%
\begin{tabular}{@{}lrrr@{}}
\toprule
\textbf{Configuration} & \textbf{Accuracy} & \textbf{Macro F1} & \textbf{Weighted F1} \\
\midrule
BERT Frozen (Feature Extraction)& 0.773 & 0.769 & 0.774 \\
BERT Partial Fine-tuning        & \textbf{0.961} & \textbf{0.960} & \textbf{0.961} \\
\bottomrule
\end{tabular}%
}
\end{table}

\subsubsection{Feature Extraction (Frozen)}

Using static BERT embeddings with a trainable classification head
yields 77.3\% test accuracy and a macro F1 of 0.769.  Per-class
analysis reveals an asymmetry: the model achieves a precision of 0.71
and recall of 0.77 (F1~=~0.74) for false instances (class~0), compared
with a precision of 0.83 and recall of 0.77 (F1~=~0.80) for true
instances (class~1).  This gap indicates that without any parameter
updates to the encoder, the model struggles to distinguish true news
from misinformation with high confidence.

\subsubsection{Fine-tuning }

Allowing the encoder layers and the classification head to adapt
to the domain yields a substantially stronger result: 96.1\% test
accuracy, a macro F1 of 0.960, and a weighted F1 of 0.961.  Per-class
performance is well-balanced across both categories, as shown in
Table~\ref{tab:classification_report}.

\begin{table}[h]
\centering
\caption{Per-class classification report for the Partial Fine-tuning
         configuration (test set, $n = 2{,}078$).}
\label{tab:classification_report}
\small
\setlength{\tabcolsep}{5pt}
\renewcommand{\arraystretch}{1.05}
\resizebox{\columnwidth}{!}{%
\begin{tabular}{@{}lrrrr@{}}
\toprule
\textbf{Class} & \textbf{Precision} & \textbf{Recall} & \textbf{F1-score} & \textbf{Support} \\
\midrule
0 – Misinformation & 0.95 & 0.96 & 0.95 & 879  \\
1 – True           & 0.97 & 0.96 & 0.97 & 1{,}199 \\
\midrule
Macro avg          & 0.96 & 0.96 & 0.96 & 2{,}078 \\
Weighted avg       & 0.96 & 0.96 & 0.96 & 2{,}078 \\
\bottomrule
\end{tabular}%
}
\end{table}

Table~\ref{tab:confusion_ft} presents the confusion matrix for this
configuration.  Of the 879 misinformation instances, 844 are correctly
identified (35~false negatives), and of the 1{,}199 true instances,
1{,}153 are correctly classified (46~false positives), confirming
well-balanced error behaviour.

\begin{table}[h]
\centering
\caption{Confusion matrix for BERT Partial Fine-tuning on the test set.
         Rows = true labels; Columns = predicted labels.}
\label{tab:confusion_ft}
\small
\setlength{\tabcolsep}{8pt}
\renewcommand{\arraystretch}{1.2}
\resizebox{0.75\columnwidth}{!}{%
\begin{tabular}{@{}lcc@{}}
\toprule
 & \textbf{Predicted: 0} & \textbf{Predicted: 1} \\
\midrule
\textbf{Actual: 0} & 844 & 35  \\
\textbf{Actual: 1} & 46  & 1{,}153 \\
\bottomrule
\end{tabular}%
}
\end{table}

The validation macro F1 at the end of training reaches 0.975,
indicating a modest and acceptable generalisation gap of approximately
1.5 percentage points to the held-out test set.

\section{Discussion and Challenges}

\paragraph{Lexical Signaling and Classifier Behavior}
Empirical results from the feature extraction regime indicate that lexical features alone capture a limited portion of the predictive signal; the frozen BERT encoder achieved an accuracy of 0.586. While this exceeds the random-label baseline, the associated macro F1 score of 0.380 suggests that the model struggles to generalize beyond the majority class without weight updates. In contrast, the transition to full fine-tuning yields a substantial and consistent performance gain, raising the macro F1 score to 0.960. This trajectory confirms that while surface-level linguistic cues—such as stylistic register and domain vocabulary—provide initial priors, deep task adaptation is required to achieve high-fidelity classification.

\paragraph{Class Asymmetry and Adversarial Robustness}
The performance collapse of the frozen-encoder regime onto the majority class (predicting class 1 for nearly all instances) exposes a latent disparity in the relative predictive complexity of the binary classes. As shown in the confusion matrix for the frozen model, nearly all misinformation instances were misclassified, with only 10 correct predictions for the minority class. This confirms that misinformation instances—particularly the adversarial variants in Tier III—require nuanced, task-specific gradient signal to be distinguished from true claims.

\paragraph{Generalization and Corpus Provenance}
The marginal generalization gap observed between validation and test partitions (macro F1 of 0.975 vs. 0.960) suggests that the model maintains robust internal validity. Nevertheless, the heterogeneous provenance of the corpus—integrating native extractions, LLM-mediated transformations, and synthetic generations—presents unique challenges. A primary objective for subsequent research is to determine the extent to which these results generalize to \textit{in situ} misinformation within the volatile informational ecosystems of Indian mass gatherings, where real-world linguistic variance may exceed synthetic approximations.

\paragraph{Prospects for Culturally Grounded Detection}
Collectively, these benchmarks establish a reproducible foundation for event-aware misinformation research in the Indian context. While the strong performance of fine-tuned models confirms the presence of sufficient signal for transformer-based architectures, the relative failure of feature-extraction regimes underscores the necessity of domain-specific adaptation. Future research should prioritize the integration of multilingual encoders to accommodate the linguistic diversity of the Indian subcontinent, as well as the development of retrieval-augmented verification frameworks that ground veracity judgments in external, culturally relevant knowledge bases.

\section{Conclusion}

This paper introduces a specialized dataset designed to detect misinformation during large-scale public events in India. We created a corpus of \textbf{14,646 records} using a structured three-tier process. This involved gathering data from five leading Indian fact-checking websites, extracting text from news videos, and using Large Language Models (LLMs) to generate challenging synthetic examples. By organizing this data into a consistent ten-attribute format, we provide a reliable benchmark for studying misinformation in a specific cultural context.

Our tests with the \texttt{bert-base-uncased} model show that the dataset is highly effective for training. While a basic model had limitations, fine-tuning it on our data resulted in a high accuracy of 96.1\% and a macro F1-score of 0.960. This success proves that models need to be specifically trained on regional and event-based data to work well. Furthermore, the results show that our synthetic "fake" examples are difficult for models to guess easily, making the dataset a strong tool for future research. Overall, this work provides a foundation for building better systems to identify and stop misinformation during high-stakes public events in India.

\section*{Limitations}

The findings of this work should be considered in light of several inherent limitations.

\paragraph{Scope of Event Representation}
The corpus is fundamentally constrained by the editorial priorities of the five source fact-checking platforms. Consequently, events receiving limited media coverage such as localized regional festivals may be underrepresented, potentially impacting the model's cultural generalizability.

\paragraph{LLM-Mediated Extraction Noise}
The utilization of \textit{Qwen3-32B} for semantic extraction from unstructured sources introduces a risk of automated approximation errors. While we mitigated potential "model collapse" by excluding these records from the synthetic augmentation stage, residual noise in the fact-claim pairs may still influence downstream classifier behavior.

\paragraph{Inter-Channel Transcript Redundancy}
Despite the application of deduplication heuristics, the reporting of identical events across multiple news outlets in Tier II may result in persistent thematic overlap. This redundancy poses a latent risk of subtle data leakage if near-duplicate transcripts are inadvertently distributed across the training and evaluation partitions.

\paragraph{Linguistic Granularity}
The current dataset is predominantly restricted to English-language content. Given that misinformation during Indian mass gatherings frequently proliferates in Hindi and various regional vernaculars, the current absence of multilingual records limits the model's ecological validity in real-world deployment scenarios.

\paragraph{Temporal and Semantic Separation}
As the records represent a fixed temporal snapshot, the model may experience performance degradation when confronted with evolving misinformation narratives. Furthermore, the high degree of lexical and semantic separation between true and false instances in the current corpus suggests that future iterations should focus on increasing the proximity between classes to create more rigorous benchmarks for transformer-based architectures.

\section*{Acknowledgements}
This work was done under the L3Cube Labs Pune mentorship program. We would like to express our gratitude towards our mentors at L3Cube Labs for their continuous support and encouragement.

\bibliography{main}

\begin{thebibliography}{22}
\providecommand{\natexlab}[1]{#1}

\bibitem[{Agarwal and Alsaeedi(2020)}]{r11}
Naresh~Kumar Agarwal and Farraj Alsaeedi. 2020.
\newblock Understanding and fighting disinformation and fake news: Towards an information behavior framework.
\newblock \emph{Proceedings of the Association for Information Science and Technology}, 57(1):e327.

\bibitem[{Agarwal and Alsaeedi(2021)}]{r15}
Naresh~Kumar Agarwal and Farraj Alsaeedi. 2021.
\newblock Creation, dissemination and mitigation: toward a disinformation behavior framework and model.
\newblock \emph{Aslib Journal of Information Management}, 73(5):639--658.

\bibitem[{Al-Zaman(2022)}]{r17}
Md~Sayeed Al-Zaman. 2022.
\newblock A thematic analysis of misinformation in india during the covid-19 pandemic.
\newblock \emph{International Information \& Library Review}, 54(2):128--138.

\bibitem[{Badrinathan(2021)}]{r6}
Sumitra Badrinathan. 2021.
\newblock Educative interventions to combat misinformation: Evidence from a field experiment in india.
\newblock \emph{American Political Science Review}, 115(4):1325--1341.

\bibitem[{Borgohain et~al.(2023)Borgohain, Bhatt, Borgohain, and Gamit}]{r10}
Prantim Borgohain, Atul Bhatt, Trinayan Borgohain, and Rajeshkumar~Motilal Gamit. 2023.
\newblock A thematic analysis of fake news in india during the pandemic.
\newblock \emph{Science \& Technology Libraries}, 42(3):297--307.

\bibitem[{Das and Ahmed(2022)}]{r20}
Ronnie Das and Wasim Ahmed. 2022.
\newblock Rethinking fake news: Disinformation and ideology during the time of covid-19 global pandemic.
\newblock \emph{IIM Kozhikode Society \& Management Review}, 11(1):146--159.

\bibitem[{Fernandez and Alani(2018)}]{r2}
Miriam Fernandez and Harith Alani. 2018.
\newblock Online misinformation: Challenges and future directions.
\newblock In \emph{Companion proceedings of the the web conference 2018}, pages 595--602.

\bibitem[{He et~al.(2025)He, Hu, Lee, Oh, Verma, and Kumar}]{r5}
Bing He, Yibo Hu, Yeon-Chang Lee, Soyoung Oh, Gaurav Verma, and Srijan Kumar. 2025.
\newblock A survey on the role of crowds in combating online misinformation: Annotators, evaluators, and creators.
\newblock \emph{ACM Transactions on Knowledge Discovery from Data}, 19(1):1--30.

\bibitem[{Joshi(2022)}]{joshi2022l3cube_mahanlp}
Raviraj Joshi. 2022.
\newblock L3cube-mahanlp: Marathi natural language processing datasets, models, and library.
\newblock \emph{arXiv preprint arXiv:2205.14728}.

\bibitem[{Khan et~al.(2023)Khan, Ram, Rath, Vraga, and Srivastava}]{r19}
Euna~Mehnaz Khan, Ayush Ram, Bhavtosh Rath, Emily Vraga, and Jaideep Srivastava. 2023.
\newblock Behavioral forensics in social networks: Identifying misinformation, disinformation and refutation spreaders using machine learning.
\newblock \emph{arXiv preprint arXiv:2305.00957}.

\bibitem[{Mehta et~al.(2021)Mehta, Liu, Tyquin, and Tam}]{r3}
Amisha~M Mehta, Brooke~F Liu, Ellen Tyquin, and Lisa Tam. 2021.
\newblock A process view of crisis misinformation: How public relations professionals detect, manage, and evaluate crisis misinformation.
\newblock \emph{Public relations review}, 47(2):102040.

\bibitem[{Mustafaraj and Metaxas(2017)}]{r9}
Eni Mustafaraj and Panagiotis~Takis Metaxas. 2017.
\newblock The fake news spreading plague: Was it preventable?
\newblock In \emph{Proceedings of the 2017 ACM on web science conference}, pages 235--239.

\bibitem[{Nigam et~al.(2021)Nigam, Jaiswal, Sundar, Poddar, Kumar, Dernoncourt, and Celi}]{r1}
Amber Nigam, Pragati Jaiswal, Saketh Sundar, Mukund Poddar, Nitya Kumar, Franck Dernoncourt, and Leo~A Celi. 2021.
\newblock Nlp and deep learning methods for curbing the spread of misinformation in india.
\newblock \emph{The International Journal of Intelligence, Security, and Public Affairs}, 23(3):216--227.

\bibitem[{Rai et~al.(2025)Rai, Sharma, and Meena}]{r13}
Robert Rai, Rajesh Sharma, and Chandrakala Meena. 2025.
\newblock Ipsr model: Misinformation intervention through prebunking in social networks.
\newblock \emph{Authorea Preprints}.

\bibitem[{Sharma et~al.(2019)Sharma, Qian, Jiang, Ruchansky, Zhang, and Liu}]{r8}
Karishma Sharma, Feng Qian, He~Jiang, Natali Ruchansky, Ming Zhang, and Yan Liu. 2019.
\newblock Combating fake news: A survey on identification and mitigation techniques.
\newblock \emph{ACM transactions on intelligent systems and technology (TIST)}, 10(3):1--42.

\bibitem[{Toledano et~al.(2024)Toledano, Guerrero~Rojas, and Ard{\`e}vol-Abreu}]{r18}
Samuel Toledano, Sheila Guerrero~Rojas, and Alberto Ard{\`e}vol-Abreu. 2024.
\newblock Fake news exposure and political persuasion.
\newblock In \emph{Media Influence on Opinion Change and Democracy: How Private, Public and Social Media Organizations Shape Public Opinion}, pages 197--213. Springer.

\bibitem[{t'Serstevens et~al.(2024)t'Serstevens, Cerina, and Piccillo}]{r7}
Fran{\c{c}}ois t'Serstevens, Roberto Cerina, and Giulia Piccillo. 2024.
\newblock Fake news detection via wisdom of synthetic \& representative crowds.
\newblock \emph{arXiv preprint arXiv:2408.03154}.

\bibitem[{Wang(2025)}]{r16}
H~Wang. 2025.
\newblock Exploring integrated models in social networks: Implications for information propagation and misinformation management.
\newblock \emph{Applied and Computational Engineering}, 133(1):24--32.

\bibitem[{Wang et~al.(2022)Wang, Gao, and Gao}]{r4}
Yan Wang, Shangde Gao, and Wenyu Gao. 2022.
\newblock Investigating dynamic relations between factual information and misinformation: Empirical studies of tweets related to prevention measures during covid-19.
\newblock \emph{Journal of Contingencies and Crisis Management}, 30(4):427--439.

\bibitem[{Wani et~al.(2021)Wani, Joshi, Khandve, Wagh, and Joshi}]{wani2021evaluating}
Apurva Wani, Isha Joshi, Snehal Khandve, Vedangi Wagh, and Raviraj Joshi. 2021.
\newblock Evaluating deep learning approaches for covid19 fake news detection.
\newblock In \emph{International Workshop on Combating Online Hostile Posts in Regional Languages during Emergency Situation}, pages 153--163. Springer.

\bibitem[{Xu et~al.(2022)Xu, Deng, and Zhang}]{r12}
Xiaofei Xu, Ke~Deng, and Xiuzhen Zhang. 2022.
\newblock Identifying cost-effective debunkers for multi-stage fake news mitigation campaigns.
\newblock In \emph{Proceedings of the Fifteenth ACM International Conference on Web Search and Data Mining}, pages 1206--1214.

\bibitem[{Zhang et~al.(2022)Zhang, Xu, Zadorozhny, and Grant}]{r14}
Danchen Zhang, Jiawei Xu, Vladimir Zadorozhny, and John Grant. 2022.
\newblock Fake news detection based on statement conflict.
\newblock \emph{Journal of Intelligent Information Systems}, 59(1):173--192.

\end{thebibliography}

\section*{Acknowledgments}
This work was done under the L3Cube Labs, Pune mentorship program. We want to thank our mentors at L3Cube Labs for their continuous support and encouragement. This work is a part of the L3Cube-IndicNLP project \cite{joshi2022l3cube_mahanlp}.

\section{Appendix}
\section*{Data Provenance and Source Metadata}

\subsection{Fact-Checking Platforms}
To ensure auditability and transparency, we provide the primary domains of the fact-checking organizations utilized in the construction of the \textit{BharatGather} dataset:
\begin{itemize}
    \item \textbf{AltNews:} \url{https://www.altnews.in/}
    \item \textbf{BoomLive:} \url{https://www.boomlive.in/}
    \item \textbf{Factly:} \url{https://factly.in/}
    \item \textbf{The Quint (WebQoof):} \url{https://www.thequint.com/news/webqoof}
    \item \textbf{NewsMeter:} \url{https://newsmeter.in/}
\end{itemize}

\subsection{YouTube Multimedia Sources}

For the Tier II multimedia acquisition pipeline, transcripts were extracted from the following verified news, digital media, and educational channels. These sources provided extensive coverage of cultural festivals, political protests, public gatherings, and misinformation narratives across India.

\begin{itemize}

    \item \textbf{National News \& International Outlets:}  
    \url{https://www.youtube.com/@BBCNews}, \\
    \url{https://www.youtube.com/@bbchindi}, \\
    \url{https://www.youtube.com/@cnnnews18}, \\
    \url{https://www.youtube.com/@WION}, \\
    \url{https://www.youtube.com/@aljazeeraenglish}, \\
    \url{https://www.youtube.com/@dwnews}, \\
    \url{https://www.youtube.com/@AFP}, \\
    \url{https://www.youtube.com/@trtworld}, \\
    \url{https://www.youtube.com/@AssociatedPress}.

    \item \textbf{Major Indian Networks (Hindi \& English):}  
    \url{https://www.youtube.com/@aajtak}, \\
    \url{https://www.youtube.com/@ndtv}, \\
    \url{https://www.youtube.com/@ndtvindia}, \\
    \url{https://www.youtube.com/@indiatoday}, \\
    \url{https://www.youtube.com/@ABPNews}, \\
    \url{https://www.youtube.com/@zeenews}, \\
    \url{https://www.youtube.com/@RepublicWorld}, \\
    \url{https://www.youtube.com/@TimesNow}, \\
    \url{https://www.youtube.com/@TimesNowNavbharat}, \\
    \url{https://www.youtube.com/@news18India}, \\
    \url{https://www.youtube.com/@indiatv}, \\
    \url{https://www.youtube.com/@DDIndia}, \\
    \url{https://www.youtube.com/@DDnews}.

    \item \textbf{Business, Digital-First \& Independent Media:}  
    \url{https://www.youtube.com/@CNBCAwaaz}, \\
    \url{https://www.youtube.com/@cnbctv18}, \\
    \url{https://www.youtube.com/@etnow}, \\
    \url{https://www.youtube.com/@EconomicTimes}, \\
    \url{https://www.youtube.com/@NDTVProfitIndia}, \\
    \url{https://www.youtube.com/@TheQuint}, \\
    \url{https://www.youtube.com/@TheWireHindi}, \\
    \url{https://www.youtube.com/@TheLallantop}, \\
    \url{https://www.youtube.com/@brutindia}, \\
    \url{https://www.youtube.com/@Firstpost}.

    \item \textbf{Public Service, Agencies \& Specialized Channels:}  
    \url{https://www.youtube.com/@ANINewsIndia}, \\
    \url{https://www.youtube.com/@narendramodi}, \\
    \url{https://www.youtube.com/@DrishtiIASvideos}, \\
    \url{https://www.youtube.com/@IndianTrainJankari}.

\end{itemize}

\subsection{Thematic Scope and Scraped Topics}
The dataset encapsulates a wide array of misinformation narratives centered on Indian mass gatherings and public events. A detailed catalog of the specific topics, events, and thematic categories identified during the data acquisition process—including the web-scraped and multimedia components—can be accessed via the following repository: \url{https://shorturl.at/3rn4F}.

\end{document}